\documentclass[conference]{IEEEtran}
\usepackage{times}
\usepackage[activate={true,nocompatibility}, final]{microtype}
\usepackage{amssymb}
\usepackage{mathrsfs}
\usepackage{prelude}
\usepackage{siunitx}
\usepackage{mathtools}
\hypersetup{colorlinks=true, linkcolor={red!50!black}, citecolor={green!80!black}, urlcolor={blue!80!black}, pdfauthor={Wil Thomason and Hadas Kress-Gazit}}

\IEEEoverridecommandlockouts
\newtheoremstyle{prettydef}%
{6pt}%
{6pt}%
{}%
{}%
{\bfseries}%
{}%
{\newline}%
{\thmname{#1}\thmnumber{ #2}:\thmnote{ \textnormal{\textit{#3}}}}

\theoremstyle{prettydef}
\newtheorem{definition}{Definition}
\newtheorem{example}{Example}

\let\vec\boldsymbol
\setlist[enumerate]{label=({\arabic*})}

\crefname{enumi}{item}{Items}
\newcounter{experiments}
\crefname{experiment}{experiment}{Experiments}

\newbool{comments}
\ifbool{comments}{\usepackage{todonotes}}{}

  \newcommand{\true}{\texttt{true}}
  \newcommand{\false}{\texttt{false}}
  \newcommand{\pset}[1]{\ensuremath{2^{#1}}}

  \newcommand{\robot}{\ensuremath{R}}

  \newcommand{\objects}{\ensuremath{\mathcal{O}}}

  \newcommand{\constraint}{\ensuremath{\phi}}
  \newcommand{\effect}{\ensuremath{\psi}}

  \newcommand{\action}{\ensuremath{\alpha}}
  \newcommand{\controller}{\ensuremath{\gamma}}
  \newcommand{\params}{\ensuremath{\Theta}}
  \newcommand{\param}{\ensuremath{\theta}}

  \newcommand{\obs}{\ensuremath{\mathcal{H}}}

  \newcommand{\state}{\ensuremath{\vec{q}}}
  \newcommand{\statespace}{\ensuremath{\mathcal{Q}}}
  \newcommand{\config}{\ensuremath{\mathcal{Q}}}
  \newcommand{\realstate}{\ensuremath{\vec{r}}}

  \newcommand{\domain}{\ensuremath{\mathcal{D}}}
  \newcommand{\sym}{\ensuremath{s}}
  \newcommand{\symbols}{\ensuremath{\mathcal{S}}}

  \tikzstyle{io} = [trapezium, trapezium left angle=70, trapezium right angle=110, text centered, draw=black, align=center]
  \tikzstyle{process} = [rectangle, text centered, draw=black, align=center]
  \tikzstyle{stepnode} = [circle, text centered, draw=black, align=center]
  \tikzstyle{decision} = [diamond, text centered, draw=black, align=center, aspect=2.5]
  \tikzstyle{arrow} = [thick, ->, >=stealth]

  \newcommand{\hkc}[1]{\ifbool{comments}{\todo[author=\textbf{HKG},color=blue!40,inline]{#1}}{}}

    \title{Counterexample-Guided Repair for Symbolic-Geometric Action Abstractions}
    \author{\authorblockN{Wil Thomason}
      \thanks{This material is based upon work supported by the Department of Defense (DoD) through the National Defense Science \& Engineering Graduate Fellowship (NDSEG) Program.
      We are grateful for this support.}
      \authorblockA{\texttt{wbthomason@cs.cornell.edu}\\
        Department of Computer Science\\
        Cornell University, Ithaca, New York 14850, USA\\
        \and
        \authorblockN{Hadas Kress-Gazit}
        \authorblockA{\texttt{hadaskg@cornell.edu}\\
          Sibley School of Mechanical and Aerospace Engineering\\
    Cornell University, Ithaca, New York 14850, USA}}}

\begin{document}

    \maketitle

    \begin{abstract}
      Integrated Task and Motion Planning (TMP) provides a promising class of approaches for solving robot planning problems with intricate symbolic and geometric constraints.
      However, the practical usefulness of TMP planners is limited by their need for \emph{symbolic abstractions} of robot actions, which are difficult to construct even for experts.
      We propose an approach to automatically construct and continuously improve a symbolic abstraction of a robot action via observations of the robot performing the action.
      This approach, called \emph{automatic abstraction repair}, allows symbolic abstractions to be initially incorrect or incomplete and converge toward a correct model over time.
      Abstraction repair uses constrained polynomial zonotopes (CPZs), an efficient non-convex set representation, to model predicates over joint symbolic and geometric state, and performs an optimizing search over symbolic edit operations to predicate formulae to improve the correspondence of a symbolic abstraction to the behavior of a physical robot controller.
      In this work, we describe the aforementioned predicate model, introduce the \emph{symbolic-geometric abstraction repair} problem, and present an anytime algorithm for automatic abstraction repair.
      We then demonstrate that abstraction repair can improve realistic action abstractions for common mobile manipulation actions from a handful of observations.
    \end{abstract}

    \IEEEpeerreviewmaketitle

    \section{Introduction}
    Integrated Task and Motion Planning (TMP), an area of robotics which seeks to holistically combine techniques from symbolic planning and geometric planning, is a promising approach to granting robots the ability to solve planning problems with intricate symbolic and geometric constraints.
    These planning problems, including many real-world mobile manipulation, assembly, and surveillance tasks, are core to practical robot utility.
    Existing work in TMP has primarily focused on \emph{efficiency} and has produced planners which can solve reasonably complex problems~\cite{garrett_sampling-based_methods_2018,dantam_incremental_task_2016,thomason_unified_sampling-based_2019,toussaint_logic-geometric_programming:_2015,srivastava_combined_task_2014}.
    However, TMP needs more than efficiency to be practically useful.
    It must also produce \emph{robust} (i.e.\ successful despite perturbations in control) solutions and be \emph{easy to use} for non-experts.
    Most current TMP techniques rely on \emph{action abstractions} (models of robot controller dynamics) which are difficult to create correctly and which have implications for planner performance.
    These abstractions are either bespoke and require expert manual effort to create, or are learned and may be difficult to verify or extend.

    As an alternative, we present a technique for automatically constructing and updating interpretable, symbolic action abstractions, called \emph{abstraction repair}.
    Abstraction repair is the problem of correcting the correspondence between an action abstraction and its underlying physical controller, guided by observations of the controller's execution.
    Solving this problem makes TMP easier to use because
    \begin{enumerate*}
      \item it allows initial (human-provided) abstractions to be simple or even incorrect, without impacting planner success,
      \item it can potentially improve planner performance by tailoring an abstraction to its use, and
      \item it can automatically adapt abstractions for different controllers and/or systems with a variety of constraints
    \end{enumerate*}.

    In this work, we
    \begin{enumerate*}
      \item adapt \emph{constrained polynomial zonotopes} (CPZs), a computationally efficient non-convex set representation with useful closure properties, as a novel set representation for symbolic-geometric planning,
      \item introduce and formally describe the \emph{symbolic-geometric abstraction repair} problem, and
      \item contribute an anytime algorithm for symbolic-geometric abstraction repair
    \end{enumerate*}.
    The abstraction representation we propose admits an automatically-derived method for sampling states that satisfy arbitrary combinations of constraints.
    Finding these states is critical to solving TMP problems and usually requires manual effort to implement.
    Our abstraction representations are also interpretable, reusable, and extensible by humans.

    \section{Related Work}
    \subsection{TMP and Action Abstractions}
    The core of most TMP work is the choice of abstraction, which bridges the symbolic and geometric components of a TMP problem.
    Though the majority of planners use STRIPS-like action descriptions (i.e.\ precondition and effect formulae over discrete predicate symbols), the symbolic-geometric abstractions used include bespoke abstraction and refinement function pairs~\cite{dantam_incremental_task_2016}, symbolic pose references with action-specific pose samplers~\cite{srivastava_combined_task_2014}, several variations on samplers for predicate-satisfying values~\cite{garrett_sampling-based_methods_2018,thomason_unified_sampling-based_2019,kaelbling_hierarchical_task_2011,dornhege_semantic_attachments_2012}, and others~\cite{garrett_ffrob:_efficient_2014,lagriffoul_efficiently_combining_2014,alami_geometrical_approach_1989,cambon_hybrid_approach_2009,barry_hierarchical_approach_2013}.
    These approaches require manual specification of the symbolic models of their actions.
    \citet{toussaint_logic-geometric_programming:_2015}, which solves TMP as a continuous optimization problem rather than abstracting geometric state into a symbolic representation, requires a similar manual specification of the transition constraints between modes corresponding to robot actions.
    In contrast, we present a hybrid abstraction method based on constrained polynomial zonotopes~\cite{kochdumper_constrained_polynomial_2020} which permits both efficient sampling of constraint-satisfying states for planning and automatic construction of symbolic models of actions.

    Work on learning models of action constraints and effects for TMP has had some recent success~\cite{xia_learning_sparse_2018,wang_active_model_2018,wang_learning_compositional_2020,konidaris_skills_symbols_2018,ames_learning_symbolic_2018}.
    These learning-based approaches tend to create probabilistic models without direct symbolic interpretation~\cite{xia_learning_sparse_2018,wang_active_model_2018,wang_learning_compositional_2020}; however, interpretability is important for allowing humans to verify and extend action abstraction models.
    Some of this line of work constructs symbolic abstractions for actions~\cite{konidaris_skills_symbols_2018,ames_learning_symbolic_2018}, but the symbols used are automatically generated and may be unintuitive for human users.
    Our work makes a different set of tradeoffs: while our abstractions do not quantify uncertainty, they have intuitive symbolic interpretations, and our chosen set representation provides state samplers for planning for free.
    Since our approach to creating models of actions is designed around modifying an incorrect starting point as needed, we are able to construct useful abstractions from fewer observations than techniques which are designed to start from no knowledge.

    Overall, this paper proposes a middle ground between the interpretable, bespoke representations of traditional TMP planners and newer learned models: a hybrid conservative approximation of action precondition and effect sets which can be symbolically manipulated, and which is interpretable, computationally efficient, and extensible.

    \subsection{Abstraction Repair}

    The symbolic planning and programming languages literatures also contribute work on plan repair~\cite{drabble_repairing_plans_1997,vanderkrogt_plan_repair_2005,bidot_plan_repair_2008,fox_plan_stability:_2006} and abstraction refinement~\cite{clarke_counterexample-guided_abstraction_2000,clarke_counterexample-guided_abstraction_2003}.
    Unlike most of the symbolic planning work on plan repair, our technique to abstraction repair works with hybrid, rather than purely symbolic, state and parameter spaces.
    The sense in which we mean ``abstraction'' (a symbolic model of a controller's constraints and effects) also differs from the sense used in most existing formal methods work on abstraction refinement.
    Some work from the hybrid systems literature~\cite{clarke_verification_hybrid_2003,alur_counter-example_guided_2003} is capable of verifying systems with hybrid state, but the goal of this work (to verify that a system satisfies a given safety property) is different from ours, and their use of ``abstraction'' again differs from ours.

    We can view the abstraction repair problem defined in~\cref{sec:repair.problem} as a model update/repair problem.
    Though this is a rich area of research~\cite{abbeel_using_inaccurate_2006,rastogi_sample-efficient_reinforcement_2018,sutton_dyna_integrated_1991,jong_model-based_function_2007,nouri_multi-resolution_exploration_2009,bernstein_adaptive-resolution_reinforcement_2010,vemula_planning_execution_2020}, we learn a different kind of model than those typically considered.
    \citet{vemula_planning_execution_2020} learns model corrections online for planning with inaccurate models, but its model updates bias the planner away from poorly modeled states rather than improving the correspondence between the model and the modeled controller.

    Finally, we draw from the literature on set representations from the formal methods community~\cite{raghuraman_set_operations_2020,scott_constrained_zonotopes:_2016,kochdumper_constrained_polynomial_2020,kochdumper_sparse_polynomial_2019,althoff_reachability_analysis_2013,le_zonotopes:_guaranteed_2013,alamo_guaranteed_state_2005} for the basis of our abstraction representation: the constrained polynomial zonotope (CPZ)~\cite{kochdumper_constrained_polynomial_2020}.
    CPZs are well-suited to TMP abstractions because they can express a broad class of non-convex sets, are closed under intersection and union operations, and are computationally and representationally efficient.

    \section{Problem}
    We introduce the following notation and definitions to formally state the symbolic-geometric abstraction repair problem.

    \subsection{Terminology and Notation}
    For a given TMP problem instance, let an \emph{object} be a physical entity in the environment, and let a \emph{symbol} be a variable with values taken from an associated domain.
    The meaning of a symbol is problem-dependent; for example, many manipulation problems use a Boolean symbol to represent whether or not the robot manipulator is empty.
    We will use \objects{} and \symbols{} to denote the sets of all objects and all symbols, respectively.
    The \emph{state space} for the problem instance is then:

    \begin{definition}[State Space]\label{def:statespace}
      \begin{equation*}\label{eqn:statespace.sig}
        \statespace = \config_R \bigtimes_{o \in \objects} \config_o \bigtimes_{\sym \in \symbols} \domain(\sym)
      \end{equation*}
      where \(\config_R\) is the configuration space of a robot \robot{}, \(\config_o\) is the configuration space of object \(o \in \objects\)
      , and $\domain(\sym)$ is the domain of symbol $\sym \in \symbols$ (i.e.\ the set of values $\sym$ can assume).
      We assume that \statespace{} is bounded in all dimensions.
      This state space construction is similar to those of~\citet{thomason_unified_sampling-based_2019} and~\citet{vega-brown_task_motion_2020}.
    \end{definition}

    In a TMP problem, an \emph{action} describes an operation a robot may perform.
    Typically, and in our use, an action abstracts a parameterized robot controller with a finite execution time (sometimes referred to as a \emph{skill}) by the states from which it can run correctly (the \emph{precondition} or \emph{constraint} of the controller) and the states which are reachable by running the controller (the \emph{effect} of the controller).
    This is the form of action abstraction used by most existing symbolic planning and TMP work.
    We define an action accordingly:

    \begin{definition}[Action]\label{def:action}
      An action \(\action_j\) abstracts a parameterized controller 
      $\controller_j$ and comprises a set of parameter vectors \(\params_j\), a \emph{constraint function} \(\constraint_j: \statespace \times \params_j \to \mathbb{B}\) capturing the conditions under which the action can be used successfully, and an \emph{effect function} \(\effect_j: \statespace \times \params_j \to \pset{\statespace}\) describing the possible states of the world after the action completes.
      \(\action_j\) is itself a function describing state transformations resulting from the robot executing the controller corresponding to \(\action_j\).
      For \(\state \in \statespace\) and \(\param \in \params_j\),
      \begin{equation*}\label{eqn:action.def}
        \action_j(\state, \param) = \left\{
          \begin{matrix}
            \effect_j(\state, \param) & \text{if } \constraint_j(\state, \param) = \true \\
            \{\state\}                & \text{otherwise}
          \end{matrix}
        \right.
      \end{equation*}
    \end{definition}

    An action $\action'_j$ is a \emph{correct abstraction} if its constraint and effect functions model the properties of $\controller_j$ exactly.
    That is, $\forall \state \in \statespace, \param \in \params_j$,
    \begin{enumerate*}
      \item $\constraint_\action(\state, \param) = \true \iff \controller_j$ runs successfully from $\state$ with parameters $\param$ and
      \item if $\constraint_\action(\state, \param) = \true$, then $\effect_\action(\state, \param)$ is exactly the set of states reachable by running $\controller_j$ from \state{} with parameters \param{}
    \end{enumerate*}.
    Similarly, an abstraction is \emph{incorrect} if one or both of these requirements is not met.
    Intuitively, these conditions mean that $\action'_j$ captures the behavior of $\controller_j$ accurately---its constraint correctly predicts whether or not $\controller_j$ will run successfully, and its effect accurately and precisely models the changes $\controller_j$ can cause on the world.

    For most realistic actions, it is impossible to verify the above: we cannot enumerate continuous parameters or state components, and some combinations of states and parameters are dangerous for some controllers.
    As such, we need a notion of correctness with respect to  observed controller executions:

    \begin{figure*}[t]
      \centering
      \begin{subfigure}[t]{.22\textwidth}
        \captionsetup{width=.95\textwidth}
        \includegraphics[width=\textwidth]{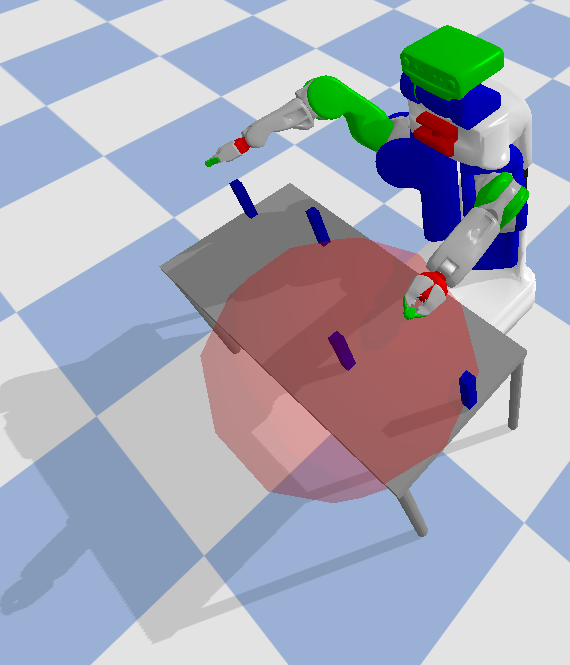}\caption{The incorrect initial abstraction of the constraint set for the \texttt{Pick} action, shown as the red sphere around the blue block.}\label{fig:example.guess}
      \end{subfigure}
      \begin{subfigure}[t]{.22\textwidth}
        \captionsetup{width=.95\textwidth}
        \includegraphics[width=\textwidth]{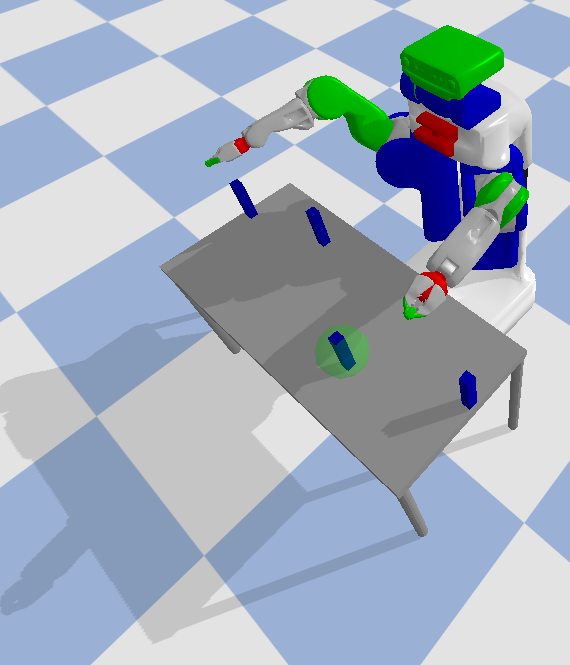}\caption{The true constraint set for the \texttt{Pick} action, shown as the green sphere around the blue block.}\label{fig:example.true}
      \end{subfigure}
      \begin{subfigure}[t]{.22\textwidth}
        \captionsetup{width=.95\textwidth}
        \includegraphics[width=\textwidth]{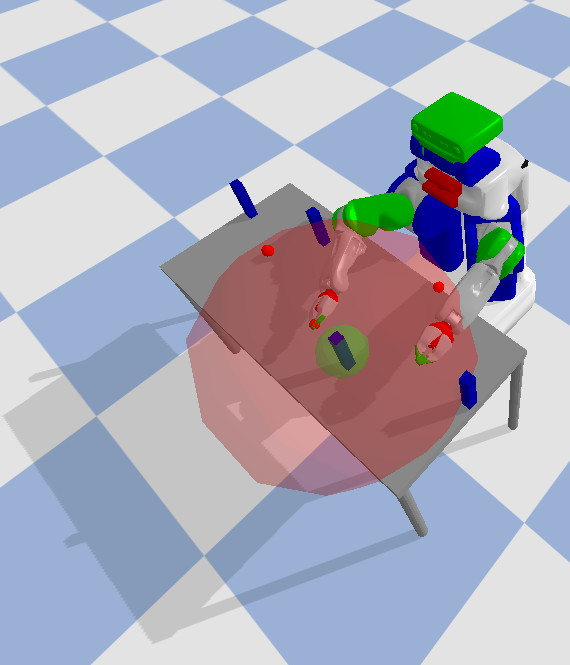}\caption{Counterexample guided abstraction repair samples states (shown as small red spheres) in the flawed abstraction set and checks if the \texttt{Pick} action succeeds.}\label{fig:example.sample}
      \end{subfigure}
      \begin{subfigure}[t]{.22\textwidth}
        \captionsetup{width=.95\textwidth}
        \includegraphics[width=\textwidth]{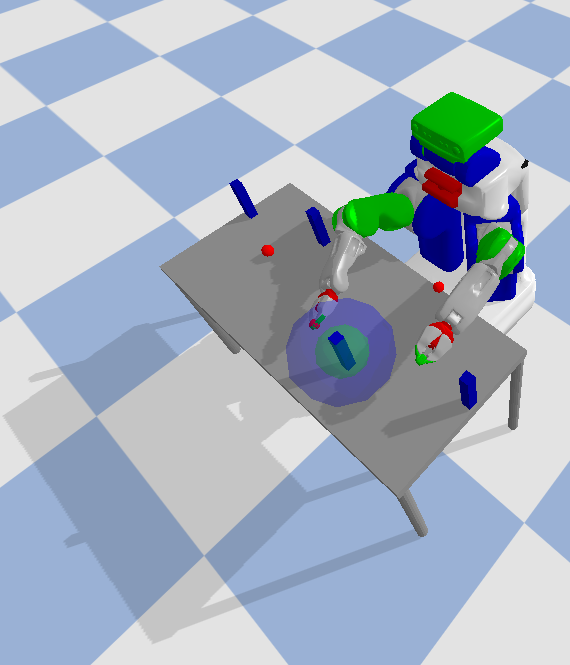}\caption{Repair uses sampled counterexamples to update the abstraction to be closer to the (unknown) true constraint set. The updated set is shown as the blue sphere.}\label{fig:example.repair}
      \end{subfigure}
      \caption{An example of the abstraction repair process for a simple \texttt{Pick} action. The goal of abstraction repair in this instance is to update a flawed initial model of an action's constraint set (\cref{fig:example.guess}) to be closer to the (unknown) true constraint set of the action (\cref{fig:example.true}), by finding counterexamples: configurations at which the model predicts a different result for the \texttt{Pick} action than is observed by running the corresponding controller (\cref{fig:example.sample}), and using these sampled counterexamples to automatically update the model of the constraint set (\cref{fig:example.repair}).}\label{fig:example}
    \end{figure*}

    \begin{definition}[Observations and Correctness]\label{def:observations}
      Let an \emph{observation} be a tuple $h_i = (\action_j, \state, \param, \state')$ recording the result $\state'$ of running $\controller_j$ from state $\state$ with parameters $\param$.
      We assume that $\action_j(\state, \param) = \{\state\}$ if and only if $\constraint_j(\state, \param) = \false$---that is, if running $\controller_j$ does not change the world, then its precondition constraint was not satisfied.

      An action $\action_j$ is then correct with respect to a set of observations \obs{} if:
      \begin{enumerate}
        \item\label{item:idempotent} $\forall h_i \in \obs$ s.t. $\state = \state'$: $\constraint_j(\state, \param) = \false$
        \item\label{item:model} $\forall h_i \in \obs$ s.t. $\state \ne \state'$: $\constraint_j(\state, \param) = \true$ and $\state' \in \effect_j(\state, \param)$
      \end{enumerate}
    \end{definition}
    Intuitively, \cref{def:observations} means that~\hyperref[item:idempotent]{(1)} for all the observations where nothing changed from running $\controller_j$, $\constraint_j$ correctly\footnote{Because $\controller_j$ is assumed to not be idempotent.} returned \false{}, and~\hyperref[item:model]{(2)} for all the observations where $\controller_j$ did change the state of the world, $\constraint_j$ correctly returned \true{} (indicating that $\controller_j$ could be successfully run) and $\effect_j$ correctly included the resulting state.

    \subsection{Symbolic-Geometric Abstraction Repair}\label{sec:repair.problem}

    We can now define the \emph{symbolic-geometric abstraction repair problem}:
    \begin{definition}[Abstraction Repair Problem]\label{def:abstraction.repair}
      Given an action $\action_j$ such that $\action_j$ is an incorrect abstraction of $\controller_j$ with respect to a set of observations \obs{}, the \emph{symbolic-geometric abstraction repair problem} is to construct an action $\action'_j$ which correctly abstracts $\controller_j$ with respect to \obs{} (per~\cref{def:observations}).
    \end{definition}

    For a simple example of an abstraction repair problem, consider the following:
    Let $\controller_p$ be a controller for pushing an object on a flat plane to a designated region.
    $\controller_p$ can be executed if
    \begin{enumerate*}
      \item the robot manipulator is empty,
      \item the robot manipulator is within 5\si{\centi\meter} of the object to be pushed, and
      \item\label{ex:path} the swept path between the object and the target region is clear
    \end{enumerate*}.
    Further let $\action_p$ be an action abstracting $\controller_p$.
    Imagine that $\constraint_p$, the constraint of $\action_p$ is incorrect, and returns \true{} whether or not the path between the object and target region is clear (it ignores (3)).
    The goal of abstraction repair in this instance, then, is to modify $\constraint_p$ to account for \hyperref[ex:path]{(3)}, given observations of the controller failing when executed with obstacles in the swept path between the target object and target region.

    As another example, which we will use as a running example, consider the task of transferring an action abstraction between two robot controllers with different preconditions.
    For instance, a PR2 may have a grasp controller which can succeed in picking up an object if it is run from a state within 10\si{\centi\meter} of the target object, regardless of pose.
    A Baxter may have a similar grasp controller, but, due to its simpler gripper, also require that the manipulator's pose be axis-aligned with the target object.
    In this instance, the goal of abstraction repair is to---starting from the known action abstraction for the PR2's controller---automatically construct a correct action abstraction for the Baxter's controller, incorporating its stricter requirements.

    \section{Modeling Constraints and Effects}
    Our approach to abstraction repair uses a new (to TMP) set representation which allows us to represent and manipulate logical formulae over sets of symbolic-geometric states.
    At a high level, we create \emph{predicate templates} underpinned by symbolic-geometric sets, which allows us to express constraint and effect functions as logical formulae.
    These formulae conservatively approximate the precondition constraints and effects of robot controllers.
    Then, upon observing a flaw in an abstraction, we apply an anytime dynamic-programming-based algorithm to ``edit'' the corresponding formula symbolically.
    This approach is amenable to using human-provided abstractions as initial input and ensures that repaired abstractions remain interpretable and reusable.
    In the following sections, we describe our set and predicate representation (\cref{sec:cpz}) and our repair algorithm (\cref{sec:algorithm}) in detail.

    \subsection{Constrained Polynomial Zonotopes}\label{sec:cpz}

    We use \emph{constrained polynomial zonotopes} (CPZs), first introduced by~\citet{kochdumper_constrained_polynomial_2020}, to represent sets of symbolic-geometric states (as defined in~\cref{def:statespace}).
    We briefly overview the structure and properties of CPZs here; the original CPZ paper offers a more detailed treatment~\cite{kochdumper_constrained_polynomial_2020}.

    A \emph{zonotope} is the Minkowski sum of a set of generating vectors, and is a commonly used set representation in the formal methods and controls communities due to its efficient representation and closure under operations such as linear map and Minkowski sum.
    Constrained polynomial zonotopes generalize zonotopes by
    \begin{enumerate*}
      \item using polynomial combinations of generating vectors and
      \item allowing for polynomial equality constraints on coefficients
    \end{enumerate*}.
    These additions grant CPZs two important properties for our use: the ability to represent many non-convex sets, and closure under intersection and union.
    CPZs also retain much of the efficient representation and computational efficiency of the simpler zonotopes.
    More formally (following~\citet{kochdumper_constrained_polynomial_2020}):

    \begin{definition}[Constrained Polynomial Zonotope]\label{def:cpz}
      An $n$-dimensional constrained polynomial zonotope $S$ is a tuple $(c, G, E, A, b, R)$ comprising a starting point $c \in \mathbb{R}^n$, a matrix of generating vectors $G \in \mathbb{R}^{n \times \ell}$, a matrix of exponent vectors $E \in \mathbb{Z}_{\ge 0}^{p \times \ell}$, a matrix of constraint generator vectors $A \in \mathbb{R}^{m \times q}$, a vector of constraint equality values $b \in \mathbb{R}^m$, and a matrix of constraint exponent vectors $R \in \mathbb{Z}^{p \times q}_{\ge 0}$.
      This tuple defines the set:
      \begin{equation*}
        \begin{split}
          S = \Biggl\{
      &c + \sum_{j = 1}^\ell \left( \prod_{k = 1}^p a_k^{E_{(k, j)}} \right) G_{(\cdot, j)} \,\Biggm|\\
      &\sum_{j = 1}^q \left( \prod_{k = 1}^p a_k^{R_{(k, j)}} \right) A_{(\cdot, j)} = b, a_k \in \left\lbrack -1, 1 \right\rbrack
    \Biggr\}
        \end{split}
      \end{equation*}
    \end{definition}

    CPZs are closed under and have closed-form exact expressions for set intersection and union~\cite{kochdumper_constrained_polynomial_2020}.

    \begin{example}[Simple CPZ]
      To build intuition for~\cref{def:cpz}, consider the following CPZ: $S_e=(c_e, G_e, E_e, A_e, b_e, R_e)$, where:
      \begin{align*}
        c_e & = \begin{bmatrix} 1 \\ 0 \end{bmatrix} &
        G_e & = \begin{bmatrix}
          2 & 1 & 2 \\
          0 & 0 & 3
        \end{bmatrix}   \\
          E_e & = \begin{bmatrix}
            1 & 0 & 1 \\
            0 & 2 & 1 \\
          \end{bmatrix} &
            R_e & = \begin{bmatrix}
              1 & 0 & 2 \\
              0 & 1 & 2
            \end{bmatrix}   \\
              A_e & = \begin{bmatrix}
                1 & 0 & 3 \\
                0 & 1 & 5 \\
                0 & 0 & 7
              \end{bmatrix} &
                b_e & = \begin{bmatrix} 2 \\ 1 \\ 2 \end{bmatrix}
              \end{align*}

              This CPZ specifies the set:
              \begin{equation*}
                \begin{split}
                  S_e = \Biggl\{
      &\begin{bmatrix} 1 \\ 0 \end{bmatrix} +
      \begin{bmatrix}
        2 \\ 0
      \end{bmatrix} a_1 +
      \begin{bmatrix}
        1 \\ 0
      \end{bmatrix} a_2^2 +
      \begin{bmatrix}
        2 \\ 3
      \end{bmatrix} a_1 a_2 \, \Biggm|\\
      &\begin{bmatrix}
        1 \\ 0 \\ 0
      \end{bmatrix} a_1 +
      \begin{bmatrix}
        0 \\ 1 \\ 0
      \end{bmatrix} a_2 +
      \begin{bmatrix}
        3 \\ 5 \\ 7
      \end{bmatrix} a_1^2 a_2^2 =
      \begin{bmatrix}
        2 \\ 1 \\ 2
      \end{bmatrix},\\
      & a_1, a_2 \in [-1, 1]
    \Biggr\}
                \end{split}
              \end{equation*}
            \end{example}

            \subsection{Encoding \statespace{} in $\mathbb{R}^n$}
            We encode a state $\state \in \statespace$ into a $\realstate \in \mathbb{R}^n$ (where \statespace{} has $n$ dimensions) for compatibility with our CPZ set representation.
            For $\state = (\state_R, \state_\objects, \state_\symbols)$, where $\state_R$ is the robot configuration vector, $\state_\objects$ is the vector of object configuration vectors, and $\state_\symbols$ is the vector of symbol values, $\realstate = (\state_R, \state_\objects, \texttt{enumerate}(\state_\symbols))$.
            $\texttt{enumerate}(\cdot)$ is a function mapping each value in the domain for each symbol to an integer.
            For example, a Boolean symbol with domain $\{\true, \false\}$ may be mapped as follows: $\{\true = 1, \false = 0\}$.
            The inverse of $\texttt{enumerate}(\cdot)$ takes a value $r_s \in \mathbb{R}$ to a value $v \in \domain(s)$ by treating the values assigned by $\texttt{enumerate}(\cdot)$ as endpoints of intervals of real numbers and determining the bin containing $r_s$.
            This encoding may result in a CPZ containing points which are not valid states (e.g.\ not all vectors of real numbers will correspond to a valid object pose in SE(3)); this shortcoming is acceptable for our use, as we are conservatively approximating sets of valid states. 

            \subsection{Modeling Constraints and Effects}
            We model constraint and effect functions as formulae over CPZs, with intersection and union taking the place of logical ``and'' and ``or'', respectively.
            We assume that logical negation is implemented at the atomic (set) level.
            This representation matches the symbolic formulae used in most existing TMP work and serves as a computationally efficient conservative approximation for arbitrary constraint and effect functions.
            Although this representation is not as general as allowing constraint and effect formulae to be arbitrary functions, it matches the representation used in most existing TMP work and suffices to represent constraints and effects for many realistic problems.
            A constraint function $\constraint_j$, then, returns true for $\constraint_j(\state, \param)$ if and only if $\state$ is contained in the CPZ described by the formula for $\constraint_j$ with parameters \param{}.
            Similarly, an effect function $\effect_j$ returns the CPZ described by the formula for $\effect_j$ from state \state{} and with parameters \param{}.

            \subsection{Predicate Templates}
            Although arbitrary CPZs can be used for these formulae, we restrict the formulae used in our repair process to a set of CPZs described by \emph{predicate templates}.
            A \emph{predicate}, for our use, is a Boolean function defining a property of a state (e.g.\ a predicate might be \true{} if a robot's manipulator is within some given distance of a particular object in a state, and \false{} otherwise---describing the property of being ``at'' the object).
            A predicate template describes a parameterized predicate, and concisely describes a whole class of properties.

            \begin{definition}[Predicate Template]\label{def:predicate.template}
              A predicate template $P_i$ pairs a \emph{constraint space transform} with a template for generating a CPZ.\@
              A constraint space transform is a function $g_i: \statespace \to \mathbb{T}_i$ (for a predicate-specific bounded $\mathbb{T}_i \subset \mathbb{R}^{m_i}$, $m_i \in \mathbb{Z}_{> 0}$), which maps states to a predicate-specific constraint space---a subspace of the real numbers in which the associated CPZ will be created.

              Constraint space transforms are related to the nonlinear constraint state maps of~\citet{chou_explaining_multi-stage_2020}, and allow us to apply arbitrary nonlinear transformations to states before describing sets as CPZs to represent a desired property.
              The rationale for using constraint space transforms is that it is often easier to describe a given state property in a transformed state space---for example, as shown in~\cref{ex:distance}, it is straightforward to create a CPZ expressing a distance constraint between a robot link and an object in the workspace, but describing this same property in the un-transformed configuration space is challenging.
              Further, we use both constraint space transforms and CPZs (rather than just having constraint space transforms describe predicates themselves) because of the compositional properties of CPZs.
              Common constraint space transforms include set-theoretic projection (for a predicate which tests only a subset of the state) and forward kinematics (for predicates best expressed in the robot's workspace).


              A template for generating a CPZ is a function $T_i$ from a predicate-specific parameter space $\params_{P_i}$ to the space of parameterized $m_i$-dimensional CPZs.
              $T_i$ constructs a parameterized CPZ representing the set of points in $\mathbb{R}^{m_i}$ (and therefore the set of states in \statespace{}) that satisfy the property described by $P_i$.
              The remaining unbound parameters in the result of $T_i$ are bound to values from the parameter vector of the constraint/effect function in which the result is used.
              For instance, in \cref{ex:distance}, the unbound parameter \texttt{obj.center} in the template result is bound to the value of the \texttt{obj} parameter for the action in which the distance predicate is used.
            \end{definition}

            Restricting constraint and effect functions to formulae over CPZs generated from a known set of predicate templates allows us to maintain interpretability when repairing an abstraction and provides useful structure to the repair search problem.
            Further, many predicate templates are applicable across a range of problems and environments, making a small set of ``basis'' predicates sufficient to describe a rich set of conditions.

            For example, we can define a predicate template to test for the value of a Boolean symbol as follows:

            \begin{example}[Symbol Value Predicate Template]
              Let $s$ be a Boolean symbol in the state space \statespace{} of a problem.
              Then the constraint space transform is $g(\state) \vcentcolon= \texttt{proj}_s(\state)$, where $\texttt{proj}_s(\cdot)$ projects a state \state{} to its value for $s$.
              The CPZ template for this predicate is the degenerate CPZ $\left( \left\lbrack 1 \right\rbrack, [], [], [], [], [] \right)$.
              Since we encode Boolean values as either 0 or 1, the above CPZ only contains states in which $s = \true$.
            \end{example}

            Similarly, a predicate template testing the distance between a robot link and an object in the environment looks like:

            \begin{example}[Distance Predicate Template]\label{ex:distance}
              This predicate template is parameterized by a distance $d$.
              The constraint space transform is the forward kinematics function $g(\state) \vcentcolon= \texttt{FK}(\state)$ for the target link.
              The CPZ template for this predicate is then $(c_{P_d}, G_{P_d}, E_{P_d}, A_{P_d}, b_{P_d}, R_{P_d})$ with
              \begingroup
              \setlength{\abovedisplayskip}{1em}
              \setlength{\belowdisplayskip}{1em}
              \setlength{\abovedisplayshortskip}{0em}
              \setlength{\belowdisplayshortskip}{0em}
              \begin{align*}
                c_{P_d} & =  \texttt{obj.center}       & b_{P_d} & =  0.5                       \\
                G_{P_d} & = dI_3                       & E_{P_d} & = \begin{bmatrix}
                  I_3 \\
                \vec{0}\end{bmatrix} \\
                  A_{P_d} & = \begin{bmatrix} 1 & 1 & 1 & -0.5 \end{bmatrix} &
                  R_{P_d} & = \begin{bmatrix}
                    2 I_3   & \vec{0} \\
                    \vec{0} & 1
                  \end{bmatrix}
                  \end{align*}
                  \endgroup
                  where $I_n$ is the $n \times n$ identity matrix.
                  This CPZ defines the ball of radius $d$ around the object \texttt{obj}, which is bound to a parameter of the action in which this predicate template is used.
                  For an object centered at $(x, y, z)$ and a particular distance value $d \in \mathbb{R}$, this produces:
                  \begin{equation*}
                    \begin{aligned}
                      \texttt{Distance}(\texttt{obj}, d) =
                      \left(\rule{0 cm}{1.1 cm}\right.
                      \begin{bmatrix} x \\ y \\ z \end{bmatrix},
                      \begin{bmatrix}
                        d & 0 & 0 \\
                        0 & d & 0 \\
                        0 & 0 & d
                      \end{bmatrix},
                      \begin{bmatrix}
                        1 & 0 & 0 \\
                        0 & 1 & 0 \\
                        0 & 0 & 1 \\
                        0 & 0 & 0
                      \end{bmatrix},      & \\
                      \begin{bmatrix} 1 & 1 & 1 & -0.5 \end{bmatrix},
                      \begin{bmatrix} 0.5 \end{bmatrix},
                      \begin{bmatrix}
                        2 & 0 & 0 & 0 \\
                        0 & 2 & 0 & 0 \\
                        0 & 0 & 2 & 0 \\
                        0 & 0 & 0 & 1
                      \end{bmatrix}
                      \left.\rule{0 cm}{1.1 cm}\right) &
                    \end{aligned}
                  \end{equation*}
                \end{example}

                \subsection{Combining CPZs with different constraint transforms}\label{sec:comining.cpzs}
                If two CPZs resulting from different predicate templates have different constraint transforms, then they must be converted to lie in a common space before we can apply intersection and/or union operations.

                The intuition behind this common space conversion is that a CPZ defined for a set of dimensions $\{d_1, \ldots, d_n\}$ does not constrain any other dimensions.
                Thus, given that each dimension of \statespace{} is bounded and that the range of each constraint space transform is bounded (from~\cref{def:statespace,def:predicate.template}, and which implies that any transformed dimension is also bounded)
                , we can re-express any pair of CPZs in a common space by adding additional generators spanning, for each original CPZ, the dimensions constrained only by the other.
                The common space CPZs are each such that
                \begin{enumerate*}
                  \item when projected down to its original dimensions, it is equivalent to the corresponding original CPZ, and
                  \item when projected down to the dimensions it did not constrain, it includes all points in those dimensions
                \end{enumerate*}.
                We include the full details of this constraint space unification procedure in~\cref{appendix:unifying}.

                \section{Abstraction Repair Algorithm}\label{sec:algorithm}
                \begin{figure*}[t]
                  \centering
                  \begin{tikzpicture}[node distance = 0.5cm, auto]
                    \node (start) {Start};
                    \node (init) [io, right=of start] {Sample initial \state{}, \param{} for $\action_j$;\\set $\action'_j=\action_j$};
                    \node (obs) [process, below=of init] {Observe controller execution\\$h_i = (\action'_j, \state, \param, \state') \in \obs$};
                    \node (unexpected) [decision, right=of obs] {$\state' \notin \effect_j(\state, \param)$?};
                    \node (repair) [stepnode, right=of unexpected, xshift=0.5cm] {Start\\repairs};
                    \node (sample) [process, below=of obs] {Sample \state{}, \param{} for $\action'_j$ \\via~\hyperref[sec:active.sampling]{active counterexample sampling}};
                    \node (precond) [process, above=of repair] {Apply~\cref{alg:repair} to $\constraint_j$ for \obs{};\\$h_i$ labeled as ``should exclude''};
                    \node (effect) [process, below=of repair] {Apply~\cref{alg:repair} to $\effect_j$ for \obs{};\\$h_i$ labeled as ``should include''};
                    \node (choose) [process, right=of repair, xshift=0.5cm] {Choose best candidate\\ edit; form $\alpha'_j$};
                    \draw [arrow] (start) -- (init);
                    \draw [arrow] (init) -- (obs);
                    \draw [arrow] (obs) -- (unexpected);
                    \draw [arrow] (unexpected) -- node[anchor=south] {Yes} (repair);
                    \draw [arrow] (unexpected.south) |-node[anchor=south east] {No} (sample.east);
                    \draw [arrow] (repair) -- (precond);
                    \draw [arrow] (repair) -- (effect);
                    \draw [arrow] (precond.east) -| node[anchor=west, align=center, yshift=-0.3cm] {Return edit\\candidate} (choose.north);
                    \draw [arrow] (effect.east) -| node[anchor=west, align=center, yshift=0.3cm] {Return edit\\candidate} (choose.south);
                    \draw [arrow] (choose.east) -- +(0.5cm, 0) -- +(0.5cm, -2.5cm) -| (sample.south);
                    \draw [arrow] (sample) -- (obs);
                  \end{tikzpicture}
                  \caption{%
                    Our abstraction repair algorithm begins by choosing an action $\action_j$ and sampling a state \state{} and parameter values \param{}.
                    We execute the controller $\controller_j(\state, \param)$ and observe the resulting state $\state'$.
                    If $\state'$ is ``expected'' (i.e.\ already contained in the set described by $\effect_j(\state, \param)$) we draw a new sample according to our~\hyperref[sec:active.sampling]{active counterexample sampling algorithm}.
                    If $\state'$ is ``unexpected'' (i.e.\ not contained in $\effect_j(\state, \param)$) we run~\cref{alg:repair} for both possible repairs (either a flaw in the constraint or in the effect for $\action_j$) until either the timeout is reached or the searches terminate.
                    We then choose the candidate edit with the lowest error, update the action abstraction accordingly, and draw a new sample according to our~\hyperref[sec:active.sampling]{active counterexample sampling algorithm}.
                    %
                  }\label{fig:approach.repair.algorithm}
                \end{figure*}
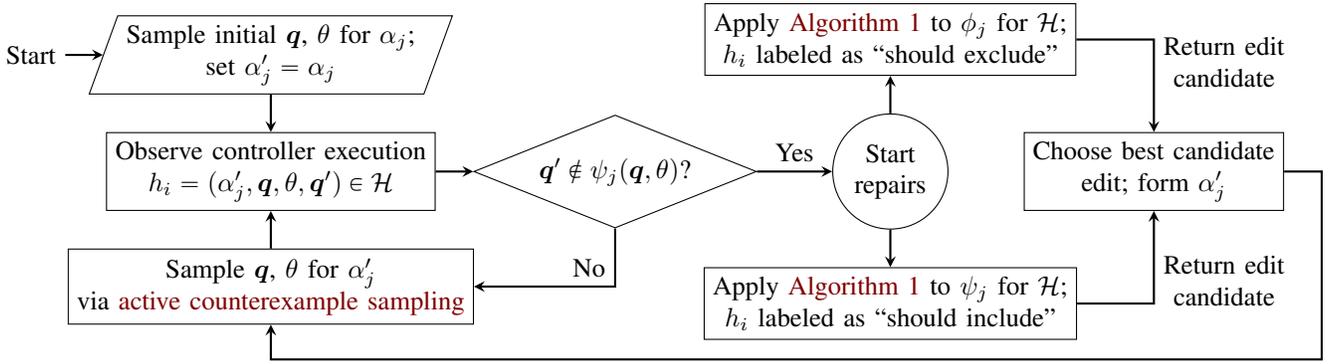

                Our approach to abstraction repair is an anytime algorithm based on symbolic ``edits'' to a constraint or effect formula.
                Intuitively, because we represent each constraint or effect as a finite-length logical formula over symbolic predicates backed by CPZs, any incorrect formula can be transformed into the correct formula for its constraint or effect by applying a finite series of edit operations which either add, remove, or replace predicates, or which modify the parameters of a predicate template.
                Further, we observe the following sound (but incomplete\footnote{This recurrence is incomplete for optimally solving abstraction repair because an intermediate edit may have a higher error and thus not be chosen.}) recurrence relation:
                \begin{align*}
                  \texttt{best}(f_{i+1}) = \min          \Bigl\{
   & \min_{\texttt{add}}(\texttt{best}(f_i)),
   \min_{\texttt{remove}}(\texttt{best}(f_i)),     \\
   & \min_{\texttt{replace}}(\texttt{best}(f_i)),
 \min_{\texttt{param}}(\texttt{best}(f_i)) \Bigr\}
                \end{align*}
                where $f_i$ is the $i^\text{th}$ edit applied to formula $f$, $\texttt{best}(f_0) = f_0$, and $\min_{\texttt{op}}(\cdot)$ computes the lowest-error single edit using \texttt{op} as its operation.
                There are several feasible candidates for the error function used in the $\min$ computations above.
                Most directly, we can measure the distance to the set boundary of observations with the wrong set inclusion value (i.e.\ if an observation should have been included in the set described by a formula but is not, or vice versa).
                This metric intuitively describes the correctness of a formula as defined in~\cref{def:observations} as well as the ``amount'' of incorrectness for misclassified observations.
                We could also favor conservative edits to a formula by using change in set measure as a component of the error function.

                We show the repair process loop in~\cref{fig:approach.repair.algorithm}.
                This process iteratively samples states and parameter values from which to attempt a given controller $\controller_j$ abstracted by an action $\action_j$, and notes whether or not the result of running $\controller_j$ is unexpected.
                If it is, the process attempts to repair both $\constraint_j$ and $\effect_j$.
                We perform repairs on both formulae because, without additional knowledge about the \emph{intended effect} of $\controller_j$, it may be impossible to distinguish between an unexpected result caused by an underconstrained constraint formula and an unexpected result caused by an overconstrained effect formula.
                The process in~\cref{fig:approach.repair.algorithm} is one possible process for repair using~\cref{alg:repair}---the core abstraction repair algorithm can also be used in the process of executing a plan, or in an interactive mode of operation with human guidance.

                We provide pseudocode for the main edit search algorithm in~\cref{alg:repair}, which implements an anytime version of the above recurrence relation.
                In~\cref{line:partition}, we extract the improperly classified observations from the set \obs{}.
                Then, in~\cref{line:init1,line:init2,line:init3}, we find the initial error value for the flawed formula $f$ and compute the initial set of candidate edits to the formula.
                We then iterate through the set of edits until either time runs out or we have evaluated every candidate edit.
                For each candidate edit, we first evaluate the edit operation to get the resulting formula (\cref{line:eval}), then compute its error with respect to the set of observations (\cref{line:error}).
                If the candidate's error is lower than or equal to the current best error, we find the new set of incorrectly classified observations and add new candidate edits to the queue (\cref{line:lequpdate,line:edits}).
                If the candidate has lower error than the current best formula, we also update the tracked best in~\cref{line:update1,line:update2,line:update3}.
                Finally, when we're out of edits or time, we return the current best formula.

                \Cref{alg:repair} relies on a few auxiliary functions: \texttt{Error} computes the error of a formula relative to a set of observations, and can be any of a variety of error functions (as discussed above).
                \texttt{Unexpected} is a Boolean function that computes whether or not an observation is correctly classified by a formula; \texttt{filter} is the standard filter function, which (in our use) returns the set of observations for which \texttt{Unexpected} returns \true{}.
                \texttt{time} returns the elapsed time since the algorithm started running.

                The \texttt{GenerateEdits} function computes the set of relevant edits for a formula and set of observations.
                It iterates through each observation in the set and finds the set of edits adding a predicate, replacing a predicate, removing a predicate, and re-optimizing the parameters of a predicate.
                Each of these four edit operations can be implemented naively, by enumerating the possible predicates, or can use information about the formula $f$ and observation $h$ to filter the set of edits returned (e.g.\ by analyzing if the observed states are correctly classified in any subset of their dimensions, and only returning predicates which apply to the remaining dimensions).
                Parameter optimization can be implemented differently for different predicates, but generally involves solving a nonlinear program minimizing \texttt{Error} over a predicate's continuous parameters, nested in an enumerative search over the domains of the predicate's discrete parameters.
                We optimize the parameters of a single predicate at a time (i.e.\ we do not explicitly solve joint parameter optimization problems).

                We additionally require that the formula $f$ is expressed in disjunctive normal form (i.e.\ a disjunctive combination of conjunctive clauses) to simplify the search space.
                This restriction reduces the number of possible sites in the formula at which to add a predicate (one site per conjunctive clause to add each predicate with logical ``and'', and one site to add each predicate as its own clause with logical ``or'' at the top level of the formula).
                %
                %
                \wtnote[inline]{Could walk through the ``running example'' from earlier here---probably worth the space.}
                \hkc{agreed - i think that will help a lot}
                \begin{algorithm}
                  \SetAlgoLined{}
                  \DontPrintSemicolon{}
                  \newcommand{\incorrect}{\ensuremath{\obs_{\texttt{I}}}}
                  \newcommand{\edits}{edits}
                  \newcommand{\result}{r}
                  \newcommand{\best}{best}
                  \newcommand{\err}{err}
                  \SetKwFunction{Error}{Error}
                  \SetKwFunction{Filter}{filter}
                  \SetKwFunction{Time}{time}
                  \SetKwFunction{Pop}{NextEdit}
                  \SetKwFunction{Push}{push}
                  \SetKwFunction{Empty}{empty}
                  \SetKwFunction{Eval}{eval}
                  \SetKwFunction{GenerateEdits}{GenerateEdits}
                  \SetKwFunction{IsUnexpected}{Unexpected}

                  \KwIn{Formula $f$; observation set \obs{}; timeout $t$}
                  \KwOut{Formula $f'$ with lower error w.r.t. \obs{}}
                  $\incorrect \leftarrow \Filter{\IsUnexpected{f}, \obs}$\;\label{line:partition}
                  $\err \leftarrow \Error{f, \obs}$\;\label{line:init1}
                  $\best \leftarrow f$\;\label{line:init2}
                  $\GenerateEdits{f, \incorrect, \edits}$\;\label{line:init3}

                  \While{$\Time{} \le t$}{%
                    \lIf{\Empty{\edits}}{%
                      \Return{\best}
                    }

                    \tcc{Lazily evaluate next edit}
                    $e \leftarrow \Pop{\edits}$\;
                    $f' \leftarrow \Eval{e}$\;\label{line:eval}
                    $\varepsilon \leftarrow \Error{f', \obs}$\;\label{line:error}
                    \If{$\varepsilon \le \err$}{\label{line:lequpdate}%
                      $\GenerateEdits{f', \incorrect, \edits}$\;\label{line:edits}
                      \If{$\varepsilon < \err$}{\label{line:update1}%
                        $\err \leftarrow \varepsilon$\;\label{line:update2}
                        $\best \leftarrow f'$\;\label{line:update3}
                      }
                    }
                  }
                  \Return{\best}
                  \caption{Anytime Abstraction Repair}\label{alg:repair}
                \end{algorithm}

                \subsection{Active Counterexample Sampling}\label{sec:active.sampling}
                \newcommand{\pstate}{\ensuremath{\vec{p}}}
                The performance of abstraction repair depends on the observations it receives.
                While \cref{alg:repair} will never make a formula less correct
                with respect to a set of observations, purely random observation sampling may not discover instances of error in a formula.
                To mitigate this issue, we introduce a form of active sampling, which heuristically samples start states and parameter values more likely to unearth a counterexample for a given formula.
                This active sampling algorithm is straightforward: after completing a repair for an action abstraction, we sample from states that satisfy the old constraint formula $\constraint$ for a state which does \emph{not} satisfy the new constraint formula $\constraint'$ (or vice versa).
                The intuition for this heuristic is that we are more likely (compared to naive sampling) to find erroneously included or excluded start states by sampling in the difference between the old and new constraint sets; this space captures the set of states which may have been missed due to over-generalization from previous samples.
                For completeness, we also return naive samples (i.e.\ sampled directly from $\constraint'$ without considering $\constraint$) with nonzero probability.

                \section{Evaluation}
                \begin{figure*}
                  \centering
                  \begin{subfigure}[t]{.49\textwidth}
                    \captionsetup{width=.95\textwidth}
                    \includegraphics[width=\textwidth]{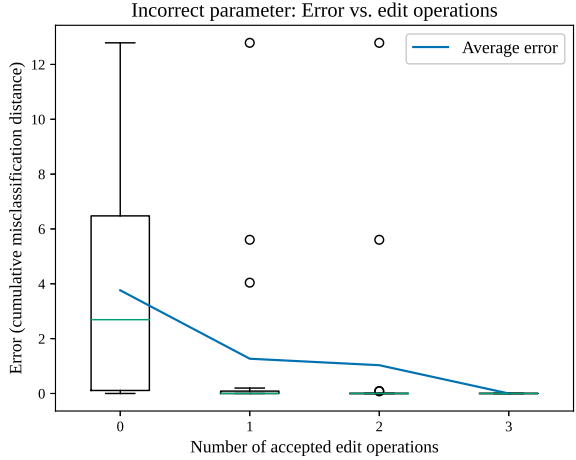}\caption{Classification error over time within a repair loop for~\cref{sec:evaluation.param}.}\label{fig:evaluation.param.classification}
                  \end{subfigure}
                  \begin{subfigure}[t]{.49\textwidth}
                    \captionsetup{width=.95\textwidth}
                    \includegraphics[width=\textwidth]{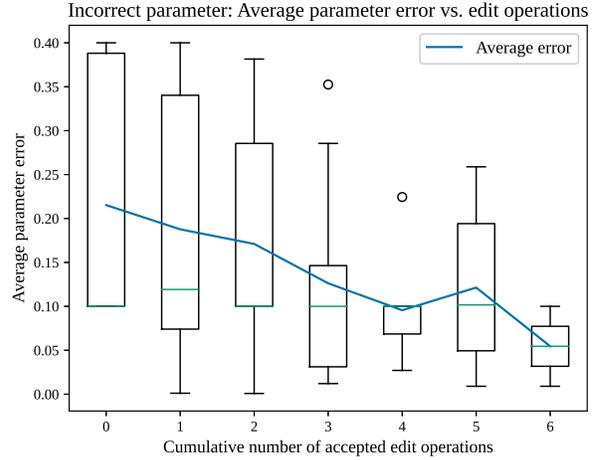}\caption{Parameter value error over a sequence of repair instances for~\cref{sec:evaluation.param}.}\label{fig:evaluation.param.value}
                  \end{subfigure}
                  \begin{subfigure}[t]{.49\textwidth}
                    \captionsetup{width=.95\textwidth}
                    \includegraphics[width=\textwidth]{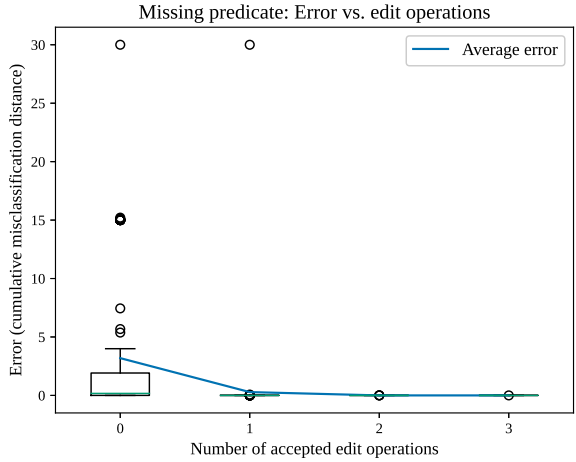}\caption{Classification error over time within a repair loop for~\cref{sec:evaluation.missing}.}\label{fig:evaluation.predicate}
                  \end{subfigure}
                  \begin{subfigure}[t]{.49\textwidth}
                    \captionsetup{width=.95\textwidth}
                    \includegraphics[width=\textwidth]{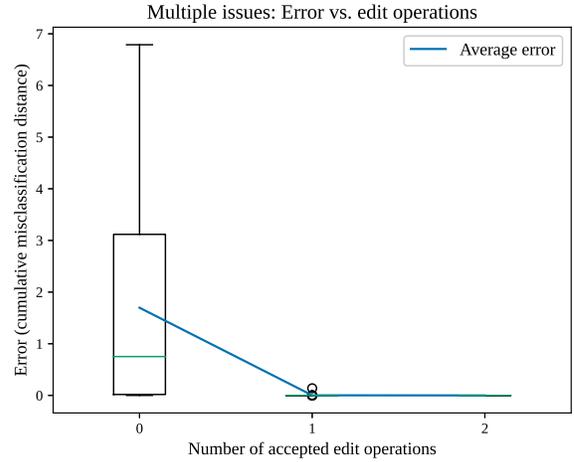}\caption{Classification error over time within a repair loop for~\cref{sec:evaluation.multiple}.}\label{fig:evaluation.multiple}
                  \end{subfigure}\caption{\Cref{fig:evaluation.param.classification,fig:evaluation.predicate,fig:evaluation.multiple} show that, within one repair loop, subsequent edit operations lead to decreased error on average.
                    There is relatively high variance in initial error (i.e.\ when zero edits have been applied but the error has been calculated), as well as several significant outliers (circles) in overall error.
                    The outliers in later steps are explained by the predicate parameter optimization sub-procedure encountering a local optimum or otherwise failing.
                    However, within at most three edit operations, all instances are able to achieve roughly zero error over their sub-sampled set of observations.
                    \Cref{fig:evaluation.param.value} shows that, for~\cref{sec:evaluation.param}, the value of the initially incorrect constraint parameter approaches the correct value as additional unexpected observations are made and repairs are performed.
                  As before, the non-monotonic variation in the parameter error is due to local optima and observation sub-sampling.}
                \end{figure*}

                We provide and evaluate an open-source Python proof-of-concept implementation of abstraction repair\footnote{Link to Github repository omitted for double-blind review.} on several pick-and-place-inspired repair operations.
                We show that we can repair, in a low number of unexpected observations\footnote{Observations for which the prediction of the abstraction under repair did not match the result of executing the corresponding controller}, realistic errors in constraint formulae\footnote{We only evaluate on constraint formula repairs; the repair process for effect formulae only changes which state in an observation is considered.}.
                Additionally, we demonstrate that abstraction repair monotonically improves the quality of abstractions (with respect to a set of observations), making it suitable for anytime use.

                All of the following experiments are single-threaded and run on a 3.7 GHz AMD Ryzen 7 2700X CPU, with 32 GB RAM.
                We use an error function computing the squared minimum distance from an observation state to the boundary of a formula's CPZ.
                This error function is smooth but not necessarily convex; we choose it because it captures the ``classification error'' of a formula (i.e.\ ``how much'' it incorrectly excludes or includes an observation).
                Given
                \begin{enumerate*}[label=(\arabic*)]
                  \item that this error function is computationally expensive (each point-CPZ distance computation requires solving a nonlinear program) and
                  \item that we want to evaluate the use of abstraction repair in an anytime context
                \end{enumerate*},
                we limit each invocation of the repair loop to a total of 100\si{s} of computation time.
                Further, we \emph{sub-sample} observations for each invocation of the repair loop.
                Given a set of unexpected observations $U$ at the time we invoke the repair loop, we select a random subset of the current expected observations $E$ such that $|U| = |E|$.
                This sub-sampling is important
                \begin{enumerate*}[label=(\arabic*)]
                  \item for computational feasibility, as computing the error function on the full set of observations is expensive, but also
                  \item to balance the expected and unexpected observations, and avoid failure cases where error is minimized by forcing a formula's CPZ to be empty (e.g.\ if most examples should be excluded)
                \end{enumerate*}.
                However, sub-sampling does have a cost: whereas~\cref{alg:repair} monotonically improves a formula given access to the full set of observations, sub-sampling can introduce variation in the error resulting from repair, leading to sub-optimal parameter values and solutions.
                We address improvements to sub-sampling in future work.
                Finally, we run repeated trials of each experiment in simulation, randomly initializing the placement of objects each time, and randomly sampling states.
                \Cref{fig:evaluation.param.classification,fig:evaluation.param.value,fig:evaluation.predicate,fig:evaluation.multiple} summarize the results of the following experiments; in these figures, outliers are shown as circles, the mean error value at each point is shown as a trend line, and the distribution of data at each point is shown as a vertical box plot.
                For~\cref{fig:evaluation.param.classification,fig:evaluation.predicate,fig:evaluation.multiple}, the $y$-axis shows error in terms of our aforementioned ``misclassification'' error function.
                For~\cref{fig:evaluation.param.value}, the $y$-axis shows the error between the actual and target value of the continuous parameter in the formula under repair.

                \subsection{Incorrect predicate parameter}\refstepcounter{experiments}\label[experiment]{sec:evaluation.param}
                As a simple example, we start by showing a repair for a single-predicate constraint formula with an incorrect continuous parameter value.
                Consider a constraint for a basic \texttt{Pick} action requiring the manipulator to be within a certain distance of the target object.
                The correct version of this formula is the single predicate $(\texttt{dist}\; obj\; manip\; 0.1)$, which says that the object $obj$ and manipulator $manip$ must be within 10\si{cm} for the grasp controller underlying the \texttt{Pick} action to successfully run.
                We start with an incorrect version of the constraint formula, $(\texttt{dist}\; obj\; manip\; 0.5)$, which is correct except for the continuous parameter value.
                Our goal, then, is to discover the correct parameter value via a series of optimization repairs in response to unexpected observations.
                We ran ten trials of this repair problem, using naive (i.e.\ non-active, uniform random) sampling for states from which to attempt the \texttt{Pick} action.
                Each trial runs until it finds five unexpected observations (and therefore invokes the repair loop five times).
                \Cref{fig:evaluation.param.classification} shows that, averaged over trials and invocations of the repair loop, we are able to decrease the formula error evaluated on the current set of observations to zero within three edit operations per loop.
                Further, \cref{fig:evaluation.param.value} shows the error in the continuous parameter value over the number of edits applied\footnote{Note that each unexpected observation causes an invocation of the repair loop, but each invocation of the repair loop may apply multiple edits.}, averaged over trials and cumulative over invocations of the repair loop.
                This trend shows that we do converge toward the correct value of the continuous parameter, although observation sub-sampling does cause the convergence to be non-monotonic.

                \subsection{Missing constraint predicate}\refstepcounter{experiments}\label[experiment]{sec:evaluation.missing}
                Next, we demonstrate that we can repair the earlier running example of transferring an action abstraction between two robot controllers with different constraints.
                We start with a constraint formula $(\texttt{dist}\; obj\; manip\; 0.1)$, which may be sufficient for a robot with a more sophisticated \texttt{Pick} controller that can align the manipulator with the object to be grasped.
                Our goal is to discover that for another robot with a simpler \texttt{Pick} controller, this formula is missing a predicate constraining the manipulator's orientation.
                Concretely, we wish to find the formula
                $
                (\texttt{and}\;  (\texttt{dist}\; obj\; manip\; 0.1)  \;
                (\texttt{roll}\; obj\; manip\; 0.1))
                $,
                where the \texttt{roll} predicate constrains the roll of the manipulator $manip$ to be within 0.1 radians of the roll of the object $obj$ in the global frame.
                In this case, uniform random sampling is too inefficient to be reasonable.
                The target formula represents a set of small measure in configuration space, so the probability of naively sampling a counterexample is similarly small and doing so takes (in expectation) a large number of wasted samples.
                Instead, we employ the active counterexample sampling algorithm described in~\cref{sec:active.sampling}, and sample directly from the repair formula's CPZ.
                We ran 20 trials of this repair problem, terminating each trial after ten unexpected observations (invoking the repair loop a total of ten times per trial).
                \Cref{fig:evaluation.predicate} shows that, as with~\cref{sec:evaluation.param}, the error (again, averaged over all trials and repair loop invocations) goes to near-zero within one to three edit operations.
                However, there is an important caveat to this result, and to this approach to abstraction repair in general: because our notion of error is defined purely by the chosen error function, it is possible for a formula a human would call spuriously correct to achieve as low an error as the correct formula, and abstraction repair has no way to distinguish which is ``better''.
                More concretely, although in some trials we discover formulae such as
                $
                (\texttt{and}\;  (\texttt{dist}\; obj\; manip\; 0.1)               \;
                (\texttt{roll}\; obj\; manip\; 0.49))
                $
                and
                $
                (\texttt{and}\;  (\texttt{dist}\; obj\; manip\; 0.08)  \;
                (\texttt{roll}\; obj\; manip\; 0.02))
                $,
                which are reasonably close in structure and parameters to the correct formula, we also discover in other trials formulae such as $(\texttt{empty}\; manip)$ and
                \begin{align*}
                  (\texttt{and}\; & (\texttt{dist}\; obj\; manip\; 4.1) \\
                                  & (\texttt{roll}\; obj\; manip\; 1.3) \\
                                  & (\texttt{empty}\; manip))
                \end{align*}
                which are clearly only spuriously correct for a particular set or sub-sample of observations (the \texttt{empty} predicate requires that the manipulator $manip$ is empty, which we model with a symbolic state component).
                Resolving this problem (and improving the performance of the search over edits) requires incorporating human knowledge (e.g.\ by presenting repair candidates with equivalent error to a human for review) or a more sophisticated selection heuristic.
                We consider this issue a primary focus for future work.

                \subsection{Multiple constraint errors}\refstepcounter{experiments}\label[experiment]{sec:evaluation.multiple}
                Finally, as both of the preceding examples have required only a single change (in the optimal case); we now show that we can repair formulae with multiple problems.
                We want to find the formula:
                \begin{align*}
                  (\texttt{and}\; & (\texttt{distance}\; obj\; manip\; 0.1) \\
                                  & (\texttt{roll}\; obj\; manip\; 0.1)     \\
                                  & (\texttt{empty}\; manip))
                \end{align*}
                starting from the formula $(\texttt{distance}\; obj\; manip\; 0.7)$, which is missing both a continuous and a discrete predicate and has an incorrect parameter value.
                As before, we ran 20 trials, stopping either after reaching 100 unexpected observations or making 1000 uninterrupted expected observations (indicating convergence).
                \cref{fig:evaluation.multiple} shows that we are able to find a formula with approximately zero error within at most two edit operations (averaged across trials and repair loop invocations); as before, we sometimes choose repaired formulae close to the desired target, and sometimes choose formulae which are only spuriously low-error.

                \section{Conclusions and Future Work}
                Integrated Task and Motion Planning, or TMP, has potential as a means of granting robots expanded autonomy, but current TMP techniques require substantial expert manual effort to use, primarily in specifying symbolic action models and samplers for constraint-satisfying states.
                The symbolic-geometric abstraction repair we propose in this work mitigates this shortcoming by allowing action models to be initially incomplete or incorrect, and therefore easier to specify.
                Coupled with our proposed novel predicate representation, based on constrained polynomial zonotopes~\cite{kochdumper_constrained_polynomial_2020}, abstraction repair can automatically produce both symbolic action models and samplers for constraint-satisfying states.
                We additionally contribute an anytime algorithm for performing abstraction repair, which monotonically improves the correctness of an action abstraction with respect to a set of observations.

                In future work, we will extend our proposed abstraction representation to
                \begin{enumerate*}
                  \item relax the need for a known set of predicate templates,
                  \item add logical negation to our CPZ logic, and
                  \item express quantified uncertainty about action constraints and effects, to better model probabilistic actions
                \end{enumerate*}.
                We will additionally improve on our algorithm for abstraction repair with regards to generalizing observations and distinguishing between solution candidates with equivalent error.
                Generalizing individual observations will improve the observation efficiency of our approach to abstraction repair, reduce the need to store a large set of observations to compute abstraction error, and permit us to more efficiently create constraint and effect formulae with universal and existential quantifiers.
                Distinguishing between edit candidates with equivalent error is a challenging problem, but could involve active learning with non-expert human users answering questions about possible repairs, or a learned heuristic approximating repair preference.

                Beyond improving upon abstraction repair, we intend to explore improvements to TMP enabled by abstraction repair and our CPZ-based predicate representation.
                Abstraction repair opens the door to online plan repair: minimally altering a plan made under a flawed abstraction to create a correct plan under the repaired abstraction.
                This would allow robots to adapt to unexpected action failures efficiently, on the fly.
                Further, CPZ-based predicate representations offer interesting possibilities for TMP planning, including estimating action difficulty by approximating precondition set measure and rapid testing for geometric feasibility via set intersection.

                \appendices
                \crefalias{section}{appendix}
                \section{Unifying Constraint Space CPZs}\label{appendix:unifying}
                As mentioned in~\cref{sec:comining.cpzs}, CPZs that lie in different spaces (resulting from different constraint transforms) must be converted to lie in a common space before we can compute their union or intersection.
                We now present the full algorithm for constructing such a common space and representing each CPZ in it.

                Let $S_1 = (c_1, G_1, E_1, A_1, b_1, R_1)$ lie in the space comprised of dimensions $D_1 = \{d_{1,1}, \ldots, d_{n,1}\}$, and let $S_2 = (c_2, G_2, E_2, A_2, b_2, R_2)$, lying in the space comprised of dimensions $D_2 = \{d_{1, 2}, \ldots, d_{m, 2}\}$.
                We construct the common space representations $S_i^c$ as follows.
                First, we must re-order the dimensions of $S_1$ and $S_2$ to be consistent.
                The common space for $S_1$ and $S_2$ is $D_c = D_1 \cup D_2$.
                We construct the ordered set $D' = \{D_1 \setminus D_2\} \cup \{D_1 \cap D_2\} \cup \{D_2 \setminus D_1\}$
                and rewrite $c_i$ and $G_i$ to $c'_i$ and $G'_i$ (respectively) by re-ordering their rows to match the order of dimensions defined by $D'$ for the dimensions in $D_i \cap D'$ (i.e.\ the dimensions constrained by $S_i$).

                Next, we form the matrix of generators for the dimensions unconstrained by $S_i$ by constructing
                \begin{equation*}
                  B_{D' \setminus D_i} =
                  \begin{bmatrix}
                    \frac{e_1}{2} &        & \vec{0}          \\
                                  & \ddots &                  \\
                    \vec{0}       &        & \frac{e_\ell}{2}
                  \end{bmatrix}
                \end{equation*}
                where $e_j$ is the extent of each dimension in $D' \setminus D_i$, in the order defined by $D'$.
                Finally, let $m_j$ be the middle point of each dimension in $D' \setminus D_i$, also in the order defined by $D'$.
                We then have:
                \begin{align*}
                  S^c_1 & = \left(
                    \begin{bmatrix}
                      c'_1   \\
                      m_1    \\
                      \vdots \\
                      m_\ell
                    \end{bmatrix},
                    \begin{bmatrix}
                      \vec{0}              & G'_1    \\
                      B_{D' \setminus D_1} & \vec{0}
                    \end{bmatrix},
                    \begin{bmatrix}
                      \vec{0} & E_1 \\ I_\ell  & \vec{0}
                    \end{bmatrix},
                    A_1,
                    b_1,
                    \begin{bmatrix}
                      R_1 \\
                      \vec{0}
                    \end{bmatrix}
                  \right)          \\
                    S^c_2 & = \left(
                      \begin{bmatrix}
                        m_1    \\
                        \vdots \\
                        m_\ell \\
                        c'_2
                      \end{bmatrix},
                      \begin{bmatrix}
                        B_{D' \setminus D_2} & \vec{0} \\
                        \vec{0}              & G'_2
                      \end{bmatrix},
                      \begin{bmatrix}
                        I_\ell  & \vec{0} \\
                        \vec{0} & E_2
                      \end{bmatrix},
                      A_2,
                      b_2,
                      \begin{bmatrix}
                        \vec{0} \\
                        R_2
                      \end{bmatrix}
                    \right)
                    \end{align*}
                    where $I_\ell$ is the $\ell \times \ell$ identity matrix and $\vec{0}$ is an appropriately-sized block of zeros.
                    Intuitively, by adding the $m_j$ to each center and the generators with $\frac{e_j}{2}$ for each missing dimension, we have added spanning vectors for the missing dimensions (because, for CPZs, the coefficients $a_i$ range from $-1$ to $1$).
                    This corresponds to the intuition above that any dimensions not present in the original CPZ should be (effectively) unconstrained in the common space representation of the CPZ.\@
                    We can then take the intersection or union (respectively) of $S^c_1$ and $S^c_2$ as usual.

                    In most practical cases, CPZs representing predicates exist in one of a few shared constraint spaces (e.g.\ pose space or a space containing a single symbol), and thus this common space transformation is often not needed.

                    \printbibliography{}
                    \end{document}